\documentclass{llncs}
\usepackage{makeidx}
\usepackage{verbatim}
\usepackage{graphicx}
\usepackage{graphicx}
\usepackage{epstopdf}
\usepackage{float}
\usepackage{subfigure}
\usepackage{amsmath}
\usepackage{booktabs}
\usepackage{multirow}
\usepackage{amssymb}

\begin{document}
\title{Learning from Suspected Target: Bootstrapping Performance for Breast Cancer Detection in Mammography}
%
%
\author{Li Xiao\inst{1}\textsuperscript{*} \and
Cheng Zhu\inst{1}\textsuperscript{*}\and Junjun Liu\inst{2}  \and
Chunlong Luo\inst{1} \and Peifang Liu\inst{2} \and  Yi Zhao\inst{1}}

%
\institute{Key Laboratory of Intelligent Information Processing, Advanced Computer Research Center, Institute of Computing Technology, Chinese Academy of Sciences, Beijing, China \\ 
\email{xiaoli@ict.ac.cn}
\and Tianjin Medical University Cancer Institute and Hospital\\
\email{juneliu2010@163.com, cjr.liupeifang@vip.163.com}}

\maketitle              
\begin{abstract}
Deep learning object detection algorithm has been widely used in medical image analysis. Currently all the object detection tasks are based on the data annotated with object classes and their bounding boxes. On the other hand, medical images such as mammography usually contain normal regions or objects that are similar to the lesion region, and may be misclassified in the testing stage if they are not taken care of. In this paper, we address such problem by introducing a novel top likelihood loss together with a new sampling procedure to select and train the suspected target regions, as well as proposing a similarity loss to further identify suspected targets from targets. Mean average precision (mAP) according to the predicted targets and specificity, sensitivity, accuracy, AUC values according to classification of patients are adopted for performance comparisons. We firstly test our proposed method on a private dense mammogram dataset. Results show that our proposed method greatly reduce the false positive rate and the specificity is increased by 0.25 on detecting mass type cancer. It is worth mention that dense breast typically has a higher risk for developing breast cancers and also are harder for cancer detection in diagnosis, and our method outperforms a reported result from performance of radiologists. Our method is also validated on the public Digital Database for Screening Mammography (DDSM) dataset, brings significant improvement on mass type cancer detection and outperforms the most state-of-the-art work.
\end{abstract}
\section{Introduction}
\begin{figure}[htbp]
\centering
\includegraphics[width=3cm, height=3cm]{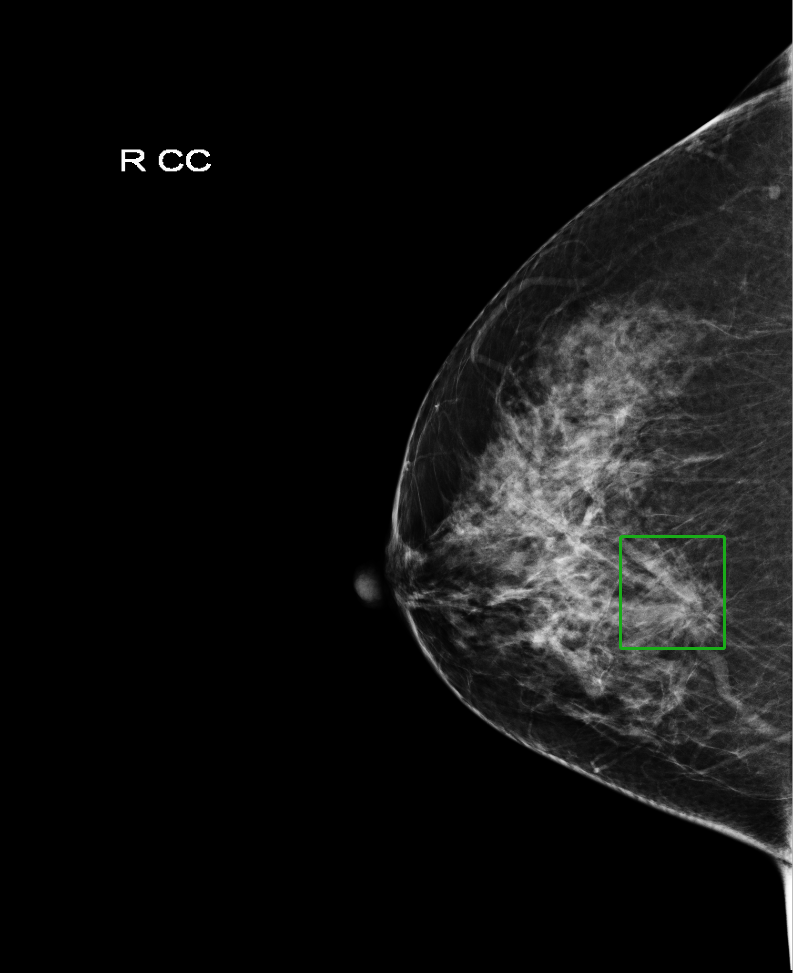}
\includegraphics[width=3cm, height=3cm]{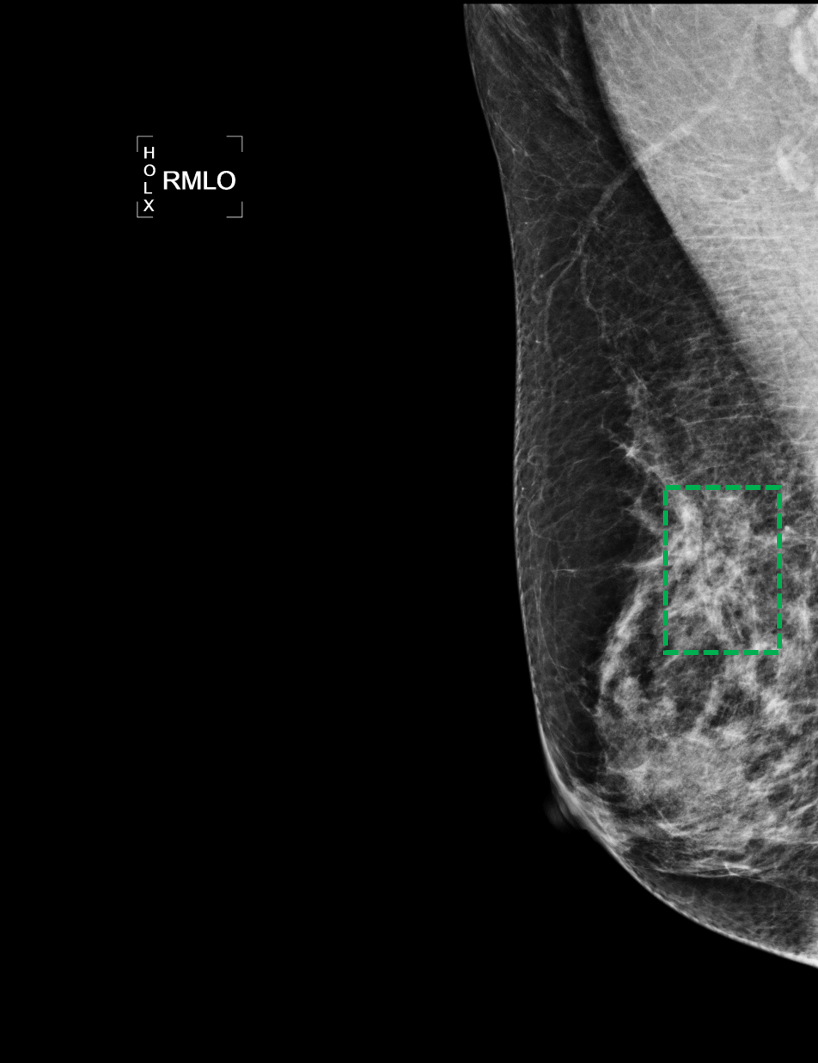}
\includegraphics[width=3cm, height=3cm]{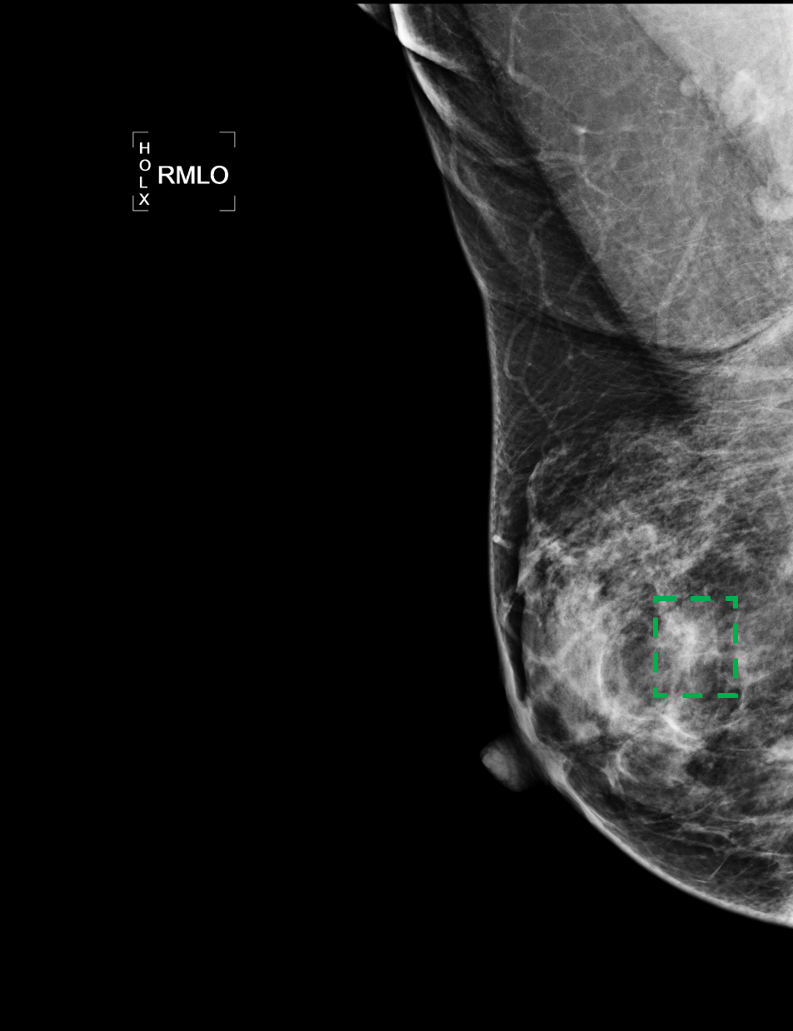}
\label{fig1}
\caption{The image with labeled malignant lesion(left) and healthy images contain unlabeled benign tumor region (middle)  or normal region(right) which are suspect to be malignant in vision.}
\vspace{-10pt}
\end{figure}

Deep learning object detection algorithm has been widely used in the task of classifying or detecting objects of natural images \cite{ref10} and \cite{ref11}, and are receiving more and more attentions on its usage in medical image analysis.
However, current object detection tasks are all based on the data annotated with object classes and their bounding boxes, those images which are not considered during labeling may contain regions or objects that are similar to the target ones, and may be misclassified in the testing stage. This phenomenon is critical in medical image analysis. For example, as shown in Fig.(\ref{fig1}), a healthy mammography may contain benign or normal regions whose features are very similar to a malignant lesion.  On the other hand, when doctors perform labeling, they usually search for medical records of patients that are diagnosed as cancer first, and only selected samples are labeled and used for training. As a result, those healthy samples that contain suspected malignant regions are likely to be classified as malignant. 

Breast cancer is one of the leading causes of death for women and mammography is the most commonly used modality for screening and early stage detection of breast cancer. Deep learning object detection models have been widely used in classification or detecting cancers of mammography\cite{dhungel2017deep,mam1,ref3,al2017detection}. 
However, current object detection models can only train well-annotated data, this will cause high false positive rate in practical usage when images with annotated malignant lesions and healthy images with no annotation are mixed together. 

In this work, we adapt the region based object detection model to a weakly-supervised learning task that encompasses both images with annotated malignant lesions and healthy images with no annotation. A top likelihood loss together with a new sampling procedure is developed to select and train those suspected target region in healthy images, and a similarity loss is further added to identify the suspected targets from targets. Our model is first validated on a private dense mammogram dataset.  By introducing the top likelihood loss and similarity loss, the false positive rate is greatly reduced and the specificity is increased by 25\%. It is worth mention that dense breast typically has a higher risk for developing breast cancer and are harder for cancer detection in diagnosis, and our method outperforms the performance of radiologists reported in Giger et.al \cite{Giger2016Automated}. Our method is then validated on the public DDSM dataset, brings significant improvement on mass type cancer detection according to all the metrics and also outperforms Al-masni et.al \cite{al2017detection}, the most state-of-the-art work. 

\section{Method}

In the following, we define a mammography with malignant lesions as a positive image, and those malignant lesions as positive targets or targets. All the other mammography are defined as negative images, including those images contain benign or normal regions that are highly suspect to be malignant(as shown in Fig.(\ref{fig1})). And without loss of generality,  we called those highly suspected malignant regions as suspected target regions. Within the datasets we used, all the positive targets are annotated with bounding boxes to indicate their precise locations and all the negative images do not have any annotation.

\subsection{Architecture}

\begin{figure}
 \centering
\includegraphics[width=0.9\textwidth]{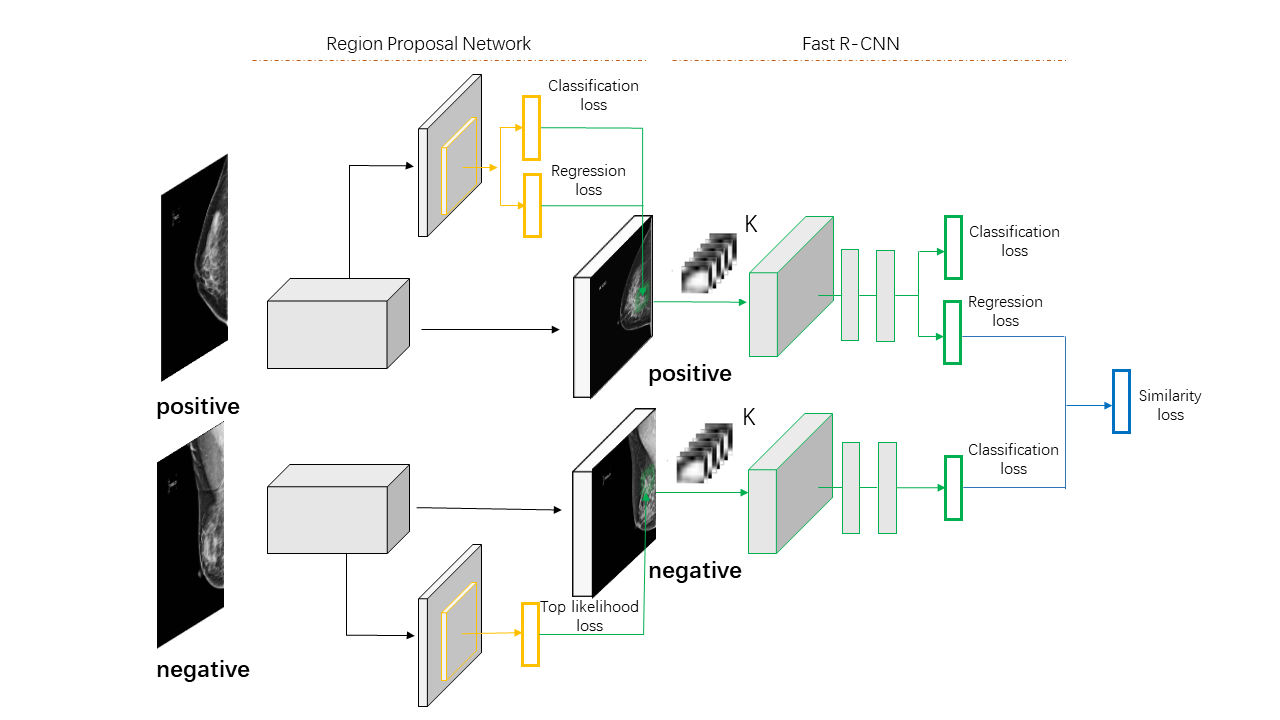}
 \caption{General architecture of our model. The input contains one positive and one negative image. In the sharing RPN stage, positive images are trained by the original loss and negative images are trained by our proposed top likelihood loss. A similarity loss  is also added on the final stage to further identify the positive and negative targets.}
 \label{fig2}
\vspace{-10pt}
\end{figure}

Our detailed network is shown in Fig.(\ref{fig2}).  The network is developed based on the Faster R-CNN \cite{fasterR-CNN}, which uses Region Proposal Network(RPN) to generate candidate proposals followed by Fast R-CNN to fine classify and regress the proposals. Both positive and negative images are randomly sorted before training. The network is modified to allow paired images as input for each mini-batch: one is positive with annotations and the other one is negative with no annotation. The positive image is trained with the original Faster R-CNN loss while the RPN loss is replaced by our newly designed top likelihood loss when training the negative image. A similarity loss is also added on the $fc$-layers of the Fast R-CNN to better identify features between positive and negative regions.

ResNet101 is adopted as the backbone. The RPN anchors are set as ($(4^{2},8^{2},16^{2},32^{2},64^{2})$) with aspect ratio as ($(\frac{1}{2},1,2)$), which cover most of the target sizes.

\begin{figure}[htbp]
\centering
\includegraphics[width=3cm, height=4cm]{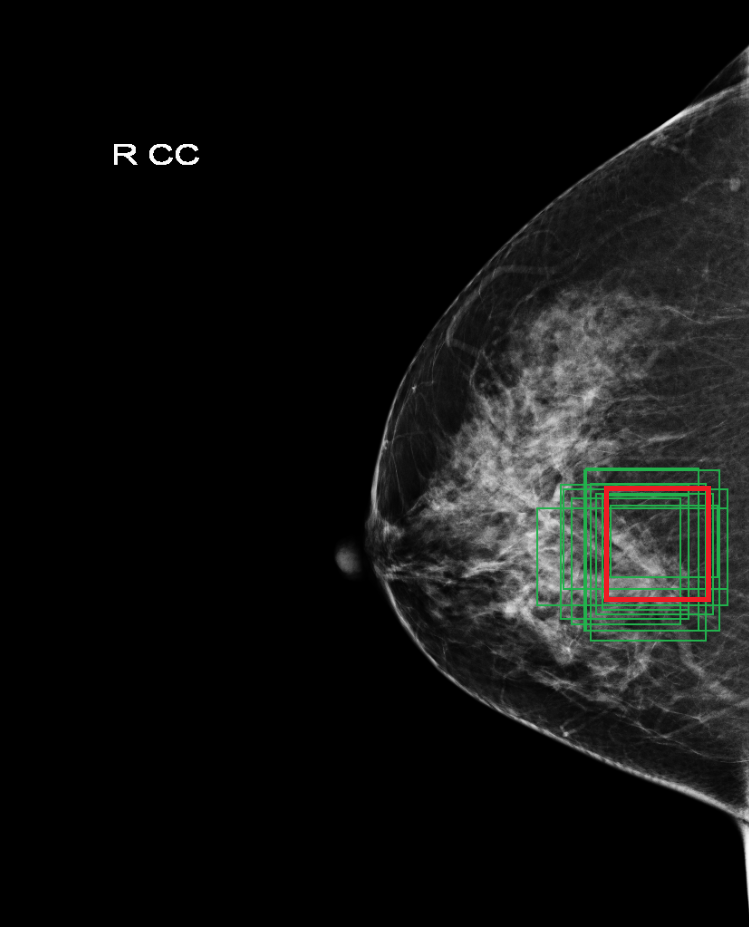}
\includegraphics[width=3cm, height=4cm]{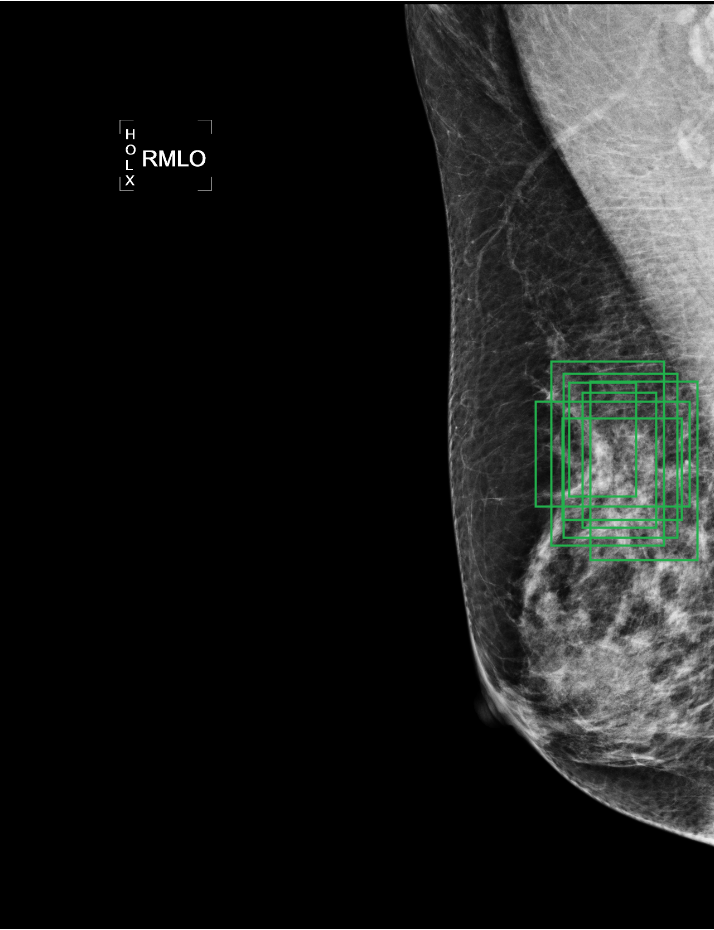}
\caption{The sampled anchors (green boxes) for positive image (left) and negative image (right) within a single mini-batch. The red box represents the ground truth target.}
\vspace{-10pt}
\label{fig3}
\end{figure}

\subsection{Loss to Discriminate Suspected Target Region}

\subsubsection{Top Likelihood Loss.}
When training the negative images, the original RPN loss is replaced by a top likelihood loss to minimize the probability to predict the negative region as positive.

In Faster R-CNN, the RPN anchor is assigned by a binary class label (of being an object or not). The positive label is assigned to two kinds of anchors: (i) the anchor/anchors with the highest Intersection-over-Union (IoU) overlap with a ground-truth box, or (ii) an anchor that has an IoU overlap higher than 0.7 with any ground-truth box. The negative label is assigned  to a non-positive anchor if its IoU ratio is lower than 0.3 for all ground-truth boxes. When training the RPN, each mini-batch randomly samples 256 anchors in an image to compute the loss function, where the sampled positive and negative anchors have a ratio of up to 1:1. If there are fewer than 128 positive samples in an image, negative ones will be padded instead.

However, the negative image does not contain any annotation, which means all anchors are negative including those anchors contain suspected targets. It is possible to optimize the loss functions of all anchors, but this will bias towards normal regions since they are dominate. Those suspected target region still likely to be recognized as positive since they don't have enough chance to be trained. To deal with this issue, we propose a top likelihood loss by ranking the scores of all anchors and sample the top 256 anchors to compute the loss function.  The anchors with top scores should be more representative for those suspected target regions as the training  goes on. On the other hand, as long as those top score anchors get minimized, all anchors are optimized towards negative at the same time.  Figure.\ref{fig3} shows the sampled anchors for the positive and negative images within a single mini-batch when training the RPN, all sampled anchors are around the suspected target region for the negative image. The top likelihood loss is defined:
\begin{equation}
L_{tlloss}=\frac{1}{cls}\sum_{i\in(tops \  p_{i})}L_{cls}(p_{i},p_{i}^{*}=0)
\label{eq1}
\end{equation}

Here, $i$ is the index of an anchor in a mini-batch and $p_{i}$ is the predicted probability of anchor $i$ being an object. Only the top 256 values of $p_{i}$ are sampled. The ground-truth label $p_{i}^{*}$ is always 0 since all anchors are negative.$L_{cls}$ loss is the cross entropy loss, same as the RPN classification loss in Faster R-CNN.

To summarize, the total RPN loss is defined as :
\begin{equation}
L_{rpnloss}=L_{ploss} + L_{nloss}= \\
        L_{pclsloss} + {\lambda_1}L_{pregloss} +{\lambda_2}L_{tlloss}
\end{equation}
Here the loss for the positive image $L_{ploss}$  is the same that in the original Faster R-CNN, which consists of the classification loss $L_{pclsloss}$ and the regression loss $L_{pregloss}$. The loss $L_{nloss}$ for the negative image only contains the top likelihood loss $L_{tlloss}$. $\lambda_1,\lambda_2$ are balancing parameters and are set to be 1 in this study.
\vspace{-15pt}
\subsubsection{Similarity Loss.}
The Fast R-CNN usually takes the top 2000 region proposals from the RPN, for each object proposal a region of interest (RoI) pooling layer extracts a fixed-length feature vector from the feature map. Each feature vector is fed into a sequence of fully connected ($fc$) layers that finally branch into two sibling output layers: one that produces softmax probability estimates the probabilities of object classes and and another layer that outputs a set of four real-valued numbers representing bounding-box positions of the classes.

An essential goal for training the negative image is to make the network better identify highly suspected malignant regions from true malignant lesions. Inspired by the successful use of Siamase loss in face recognition \cite{ref7}, which minimizes a discriminative loss function that drives the similarity metric to be small for pairs of faces from the same person, and large for pairs from different persons.  We introduce a similarity loss applied on the softmax probability layer of the classification branch in the Fast R-CNN, aiming at improving the network's ability to discriminate the positive region from negative ones. Specifically, during each mini-batch, we selected the feature vectors of the softmax probability in the classification branch whose label is 1(target label as the original Faster R-CNN defines) when the network processes the positive image. Those features vectors will then maintain the explicit positive features, we denote the number of the feature vectors as $K$. When the network processes the negative image, we take the same number of feature vectors generated by the top $K$ score anchors from RPN, and pair them with the selected feature vectors of positive images.

A similarity loss is then obtained from the $K$ paired feature vectors as:
\begin{equation}
L_{simloss}(X_1,X_2)=\frac{1}{K}\sum_{i=1}^{K}sim(X_1^i, X_2^i)
\end{equation}
Where $X_1$ and $X_2$ are the feature vectors selected from positive and negative images. $sim$ is  the cos-embedding function where $sim(X_1^i, X_2^i)=\frac{{X_1^i}\cdot{X_2^i}}{\Vert{X_1^i}\rVert\Vert{X_2^i}\rVert}$. The loss is minimized during training to increase the network's discriminability between the positive and selected negative targets.

The negative image shares the same Fast R-CNN loss as the positive image, but since there is no target, the ground-truth label  of all the proposals are 0 and no regression loss occurs. The total Fast R-CNN loss is then summarized as
\begin{equation}
L_{fastloss}=L_{clsloss} + {\lambda_3}L_{regloss} +{\lambda_4}L_{simloss}~,
\end{equation}
which consists of three terms: the classification loss $L_{clsloss}$, the regression loss $L_{regloss}$, and the similarity loss $\L_{simloss}$.  $\lambda_3,\lambda_4$ are balancing parameters and are set to be 1 and 0.1 in this study.
\vspace{-10pt}
\subsection{Datasets and Evaluation Metrics}
\vspace{-5pt}
\subsubsection{Datasets.}To test the effectiveness of our proposed top likelihood loss and similarity loss, we adopted two mammography datasets for evaluation.

\textbf{(1) Private Dense Mammogram Dataset.} a private dense mammogram dataset is collected and labeled from the Tianjin Medical University Cancer Institute and Hospital, which consists 721 patients including 417 heathy ones and 304 sick ones. Mammography includes the cranial cardo (CC) and media lateral oblique (MLO) views for most of the screened breasts. Each patient usually have 4 mammography images (R-CC,R-MLO,L-CC, L-MLO), but a few of them only have unilateral mammography images. A few of the lesions are not labeled due to visual artifacts obscuring the image data, including paddles within magnification views or location markers. There are totally 2908 images with 598 labeled malignant lesions. Noticeably, dense breast has a higher risk in developing breast cancer and are harder for cancer detection based on mammography (AUC of 0.72 according to Giger et.al \cite{Giger2016Automated}). Only mammographies with non-specific invasive carcinoma is collected and annotated. The dataset was randomly split to the training and testing set with a ratio of 4:1.

\textbf{(2) DDSM.} The public DDSM dataset\cite{DDSM} is also used to validate our model. Following the work in \cite{al2017detection}, we selected a set of 600 mammograms of the mass type from DDSM which includes 304 malignant cases and 296 benign cases as the training set. The testing set contains 45 malignant cases and 55 benign cases. All annotations of the benign cases are removed to generate unlabeled negative images.
\vspace{-15pt}
\subsubsection{Evaluation Metrics.}Two evaluation metrics are introduced to measure the performances of our model. The first one focus on the performance according to object detection, in which the commonly used mean average precision(mAP) are adopted and measured for the predicted targets. True positive IoU matching threshold is set as 0.5 which is the standard criterion in the PASCAL VOC \cite{ref10} competition. The second one is based on the clinical criterion, in which the sensitivity, specificity, overall accuracy followed by a ROC curve are adopted and measured according to classification of patients. The predicted box with a confidence probability above 0.5 is treated as a predicted target. The sensitivity, specificity and overall accuracy are defined as:
\begin{equation}
\begin{aligned}
&Sensitivity = \frac{TP}{(TP+FN)},\ \ Specificity = \frac{TN}{(TN+FP)} \\
&Accuracy = \frac{(TP+TN)}{(TP+TN+FP+FN)}
\end{aligned}
\end{equation}
where TP and FN denote the true positive and false negative patient cases. TN and FP represent the true negative and false positive patient cases.

\section{Results and Discussion}

\subsubsection{Training Details.}  All images are resized to 512*512, the initialization settings of the network are the same as that in the original Faster R-CNN\cite{fasterR-CNN}. Each mini-batch contains one positive and one negative image. We employ the Adam optimization method. Learning rate is set to be 0.0001 and is reduced by a factor of 10 every 9k iterations. The overall training procedure is continued for up to 45k iterations.

\subsubsection{Ablation Study on Private Dataset.}
\begin{figure}
\centering
\includegraphics[height=4cm,width=8cm]{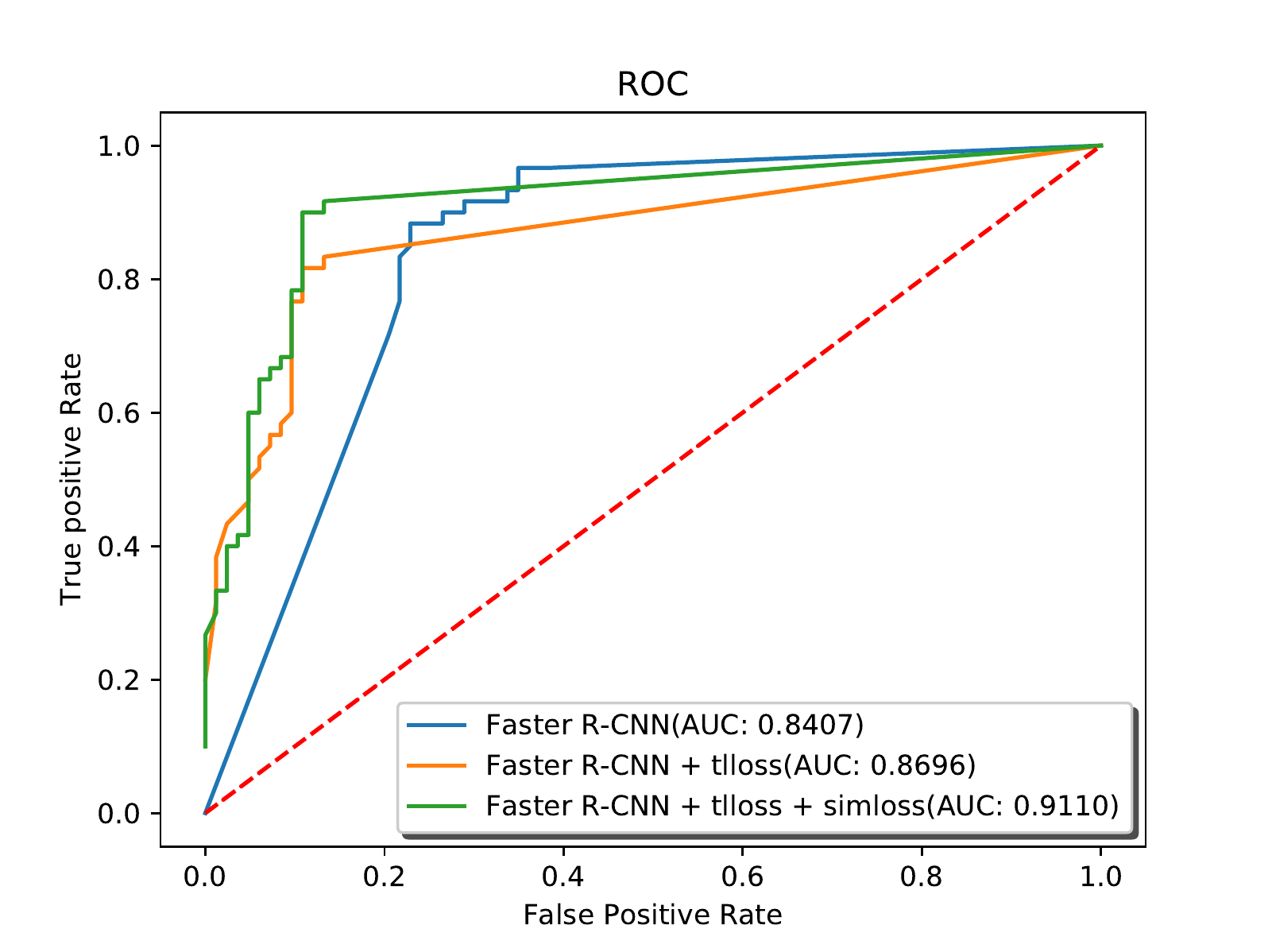}
\caption{The ROC curve of our proposed method on the private dataset. Both top likelihood loss(tlloss) and similarity loss(simloss) improve the performance. }
\label{auc}
\end{figure}

Ablation study were performed on the private mammography set with results summarized in Table \ref{tab:tab2}. The ROC curve is plotted in Figure \ref{auc}. Both top likelihood loss and similarity loss significantly improves the performances for almost all the metrics. The AUC of the classification can be as high as 0.91, which greatly outperform the AUC(0.72) of radiologists' performance reported in Giger et.al \cite{Giger2016Automated}. It is worth mention that the original Faster R-CNN which does not train negative images, results in a low specificity since the high false positive rate caused by misclassifying suspected targets. The specificity is increased by 0.25 after introducing the top likelihood loss. It is also interesting to notice that training on the negative images degrades the performance on true positive targets at some extent, but the similarity loss makes positive targets more distinguishable and improves the sensitivity.

\begin{table}[htbp]
    \centering
    \begin{tabular}{c|c|c|c|cccc}
      \hline
        & tlloss & simloss & mAP &AUC(\%) &accuracy(\%)  &sensitivity(\%)  &specificity(\%)  \\
        \hline
        \multirow{3}*{Faster R-CNN}& &  &0.52 &84.07 &76.22 &96.67 &61.45 \\
        \cline{2-8}
        &\checkmark &  &0.57 &86.96 &85.31 &83.33 &86.75\\
        \cline{2-8}
        &\checkmark &\checkmark  &0.60 &91.10 &88.81 &91.67 &86.75\\
        \hline
    \end{tabular}
\caption{ Performance of our proposed method on the private dense mammograpm dataset.}
\label{tab:tab2}
\end{table}

\subsubsection{Results and Comparisons on DDSM.} The performance on the public DDSM dataset is shown in Table \ref{tab:tab3}. We also train the Faster R-CNN with annotated benign cases for comparison. Results show that our proposed model greatly improve the performance according to all the metrics compare to baseline. Especially, the specificity is increased from 0.76 to 0.95. The performance is much better than that in AI-masni et.al \cite{al2017detection}, the most state-of-the-art work on mass type cancer detection with the same size of training set. Surprisingly, our model also outperforms the baseline even when the annotations of the benign cases are added back for training, with sensitivity increased from 0.87 to 0.91. The increased sensitivity may attribute to that training suspected malignant region makes malignant regions more distinguishable.

\begin{table}[htbp]
\begin{small}
    \centering
    \begin{tabular}{c|cccc}
      \hline
        &AUC(\%) &accuracy(\%)  &sensitivity(\%)  &specificity(\%)  \\
        \hline
        Faster R-CNN(no annotation on benign) & 89.45 &81.00 &86.67 &76.36 \\
        \hline
        Faster R-CNN+tlloss+simloss &94.46  &93.00 &91.11 &94.55 \\
        \hline
        Faster R-CNN(with annotated benign) &92.69 &91.00 &86.67 &94.55 \\
           \hline
        al-masni et.al &87.74 &85.52 &93.20 &78.00 \\
        \hline
    \end{tabular}
\caption{Performance of our proposed method on the public dataset(DDSM). }
\label{tab:tab3}
\end{small}
\vspace{-20pt}
\end{table}
\section{Conclusion}

We adapted the region based object detection model to a weakly-supervised learning task that encompasses both images with annotated malignant lesions and healthy images with no annotation. A top likelihood loss together with a new sampling procedure and a similarity loss is distinguish the suspected target from target. The new method demonstrates significant performance improvement in detecting mass type cancer on a private dense mammogram dataset and the public DDSM dataset. We would like to extend this model to other object detection tasks in which normal images contain highly suspected target regions. We would also like to apply the model to detect other type of mammographically visible lesions such as calcifications and architectural distortions.

\section{Acknowledgements}
This work was supported by the CAS Pioneer Hundred Talents Program (2017-074) to Li Xiao.

\bibliographystyle{splncs04}
\bibliography{zc}

\begin{thebibliography}{10}
\providecommand{\url}[1]{\texttt{#1}}
\providecommand{\urlprefix}{URL }
\providecommand{\doi}[1]{https://doi.org/#1}

\bibitem{al2017detection}
Al-masni, M.A., Al-antari, M.A., Park, J., Gi, G., Kim, T.Y., Rivera, P.,
  Valarezo, E., Han, S.M., Kim, T.S.: Detection and classification of the
  breast abnormalities in digital mammograms via regional convolutional neural
  network. In: 2017 39th Annual International Conference of the IEEE
  Engineering in Medicine and Biology Society (EMBC). pp. 1230--1233. IEEE
  (2017)

\bibitem{dhungel2017deep}
Dhungel, N., Carneiro, G., Bradley, A.P.: A deep learning approach for the
  analysis of masses in mammograms with minimal user intervention. Medical
  image analysis  \textbf{37},  114--128 (2017)

\bibitem{ref10}
Everingham, M., Van~Gool, L., Williams, C.K., Winn, J., Zisserman, A.: The
  pascal visual object classes (voc) challenge. International journal of
  computer vision  \textbf{88}(2),  303--338 (2010)

\bibitem{Giger2016Automated}
Giger, M.L., Inciardi, M.F., Edwards, A., Papaioannou, J., Drukker, K., Jiang,
  Y., Brem, R., Brown, J.B.: Automated breast ultrasound in breast cancer
  screening of women with dense breasts: Reader study of mammography-negative
  and mammography-positive cancers. Ajr American Journal of Roentgenology
  \textbf{206}(6), ~1 (2016)

\bibitem{ref7}
Hadsell, R., Chopra, S., LeCun, Y.: Learning a similarity metric
  discriminatively, with application to face verification. In: 2005 IEEE
  Computer Society Conference on Computer Vision and Pattern Recognition
  (CVPR'05)(CVPR). vol.~01, pp. 539--546 (06 2005).
  \doi{10.1109/CVPR.2005.202},
  \url{doi.ieeecomputersociety.org/10.1109/CVPR.2005.202}

\bibitem{DDSM}
Heath, M., Bowyer, K., Kopans, D., Moore, R., Kegelmeyer, P.: The Digital
  Database for Screening Mammography (2001)

\bibitem{mam1}
Kooi, T., Litjens, G., Van~Ginneken, B., Gubern-M{\'e}rida, A., S{\'a}nchez,
  C.I., Mann, R., den Heeten, A., Karssemeijer, N.: Large scale deep learning
  for computer aided detection of mammographic lesions. Medical image analysis
  \textbf{35},  303--312 (2017)

\bibitem{ref11}
Lin, T.Y., Maire, M., Belongie, S., Hays, J., Perona, P., Ramanan, D.,
  Doll{\'a}r, P., Zitnick, C.L.: Microsoft coco: Common objects in context. In:
  European conference on computer vision. pp. 740--755. Springer (2014)

\bibitem{ref3}
Reiazi, R., Paydar, R., Abbasian~Ardakani, A., Etedadialiabadi, M.: Mammography
  lesion detection using faster r-cnn detector. pp. 111--115 (01 2018).
  \doi{10.5121/csit.2018.80212}

\bibitem{fasterR-CNN}
Ren, S., He, K., Girshick, R., Sun, J.: Faster r-cnn: Towards real-time object
  detection with region proposal networks. In: Advances in neural information
  processing systems. pp. 91--99 (2015)

\end{thebibliography}

\end{document}